\documentclass{article}

\usepackage[nonatbib,final]{neurips_2025}

\usepackage[utf8]{inputenc} % allow utf-8 input
\usepackage[T1]{fontenc}    % use 8-bit T1 fonts
\usepackage{hyperref}       % hyperlinks
\usepackage{url}            % simple URL typesetting
\usepackage{booktabs}       % professional-quality tables
\usepackage{amsfonts}       % blackboard math symbols
\usepackage{nicefrac}       % compact symbols for 1/2, etc.
\usepackage{microtype}      % microtypography
\usepackage[table]{xcolor}
\usepackage{amsmath}
\usepackage{graphicx}
\usepackage{multirow}
\usepackage{wrapfig}
\usepackage{makecell}

\definecolor{cityblue}{RGB}{128, 159, 225}
\definecolor{citypink}{RGB}{227, 108, 194}

\definecolor{colorbest}{RGB}{252,187,161}
\definecolor{colorsecond}{RGB}{254,224,210}
\definecolor{colorthird}{RGB}{255,245,240}
\newcommand{\first}[0]{\cellcolor{colorbest} }
\newcommand{\second}[0]{\cellcolor{colorsecond}}
\newcommand{\third}[0]{\cellcolor{colorthird}}
\DeclareRobustCommand{\legendsquare}[1]{%
  \textcolor{#1}{\rule{2ex}{2ex}}%
}
\title{Step1X-3D: Towards High-Fidelity and Controllable Generation of Textured 3D Assets}

\author{
  Step1X-3D Team \& LightIllusions Team\\
  \texttt{StepFun} \& \texttt{LightIllusions} \\
  {\href{https://github.com/stepfun-ai/Step1X-3D}{\color{citypink}\textbf{\texttt{https://github.com/stepfun-ai/Step1X-3D}}}}
}

\begin{document}

\newcommand{\wy}[1]{{\color{orange}[#1]}}
\newcommand{\wyc}[1]{{\color{orange}[wy: #1]}}
\newcommand{\cw}[1]{{\color{cityblue}[cw: #1]}}

\maketitle

\begin{abstract}
While generative artificial intelligence has advanced significantly across text, image, audio, and video domains, 3D generation remains comparatively underdeveloped due to fundamental challenges such as data scarcity, algorithmic limitations, and ecosystem fragmentation. 
To this end, we present Step1X-3D, an open framework addressing these challenges through: 
(1) a rigorous data curation pipeline processing >5M assets to create a 2M high-quality dataset with standardized geometric and textural properties; 
(2) a two-stage 3D-native architecture combining a hybrid VAE-DiT geometry generator 
with an diffusion-based texture synthesis module; and (3) the full open-source release of models, training code, and adaptation modules. For geometry generation, the hybrid VAE-DiT component produces TSDF representations by employing perceiver-based latent encoding with sharp edge sampling for detail preservation. The diffusion-based texture synthesis module then ensures cross-view consistency through geometric conditioning and latent-space synchronization.
Benchmark results demonstrate state-of-the-art performance that exceeds existing open-source methods, while also achieving competitive quality with proprietary solutions. 
Notably, the framework uniquely bridges the 2D and 3D generation paradigms by supporting direct transfer of 2D control techniques~(e.g., LoRA) to 3D synthesis.
By simultaneously advancing data quality, algorithmic fidelity, and reproducibility, Step1X-3D aims to establish new standards for open research in controllable 3D asset generation.
\end{abstract}

% \section{Approach}

\begin{figure*}[h!]
  \centering
  \includegraphics[width=\linewidth]{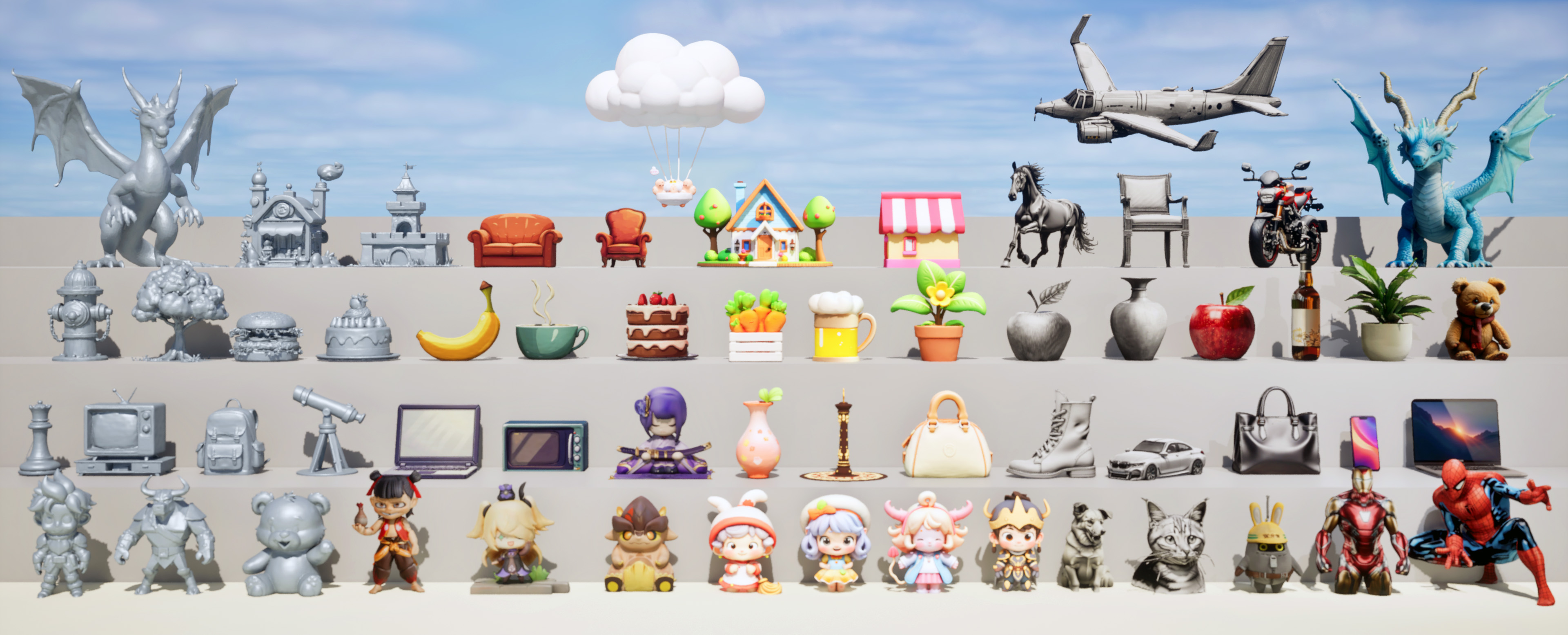}
  \caption{Step1X-3D demonstrates the capability to generate 3D assets with high-fidelity geometry and versatile texture maps, while maintaining exceptional alignment between surface geometry and texture mapping. From left to right, we sequentially present: the base geometry~(untextured), followed by cartoon-style, sketch-style, and photorealistic 3D asset generation results.}
  \label{fig:teaser}
\end{figure*}

\section{Introduction}
In recent years, generative artificial intelligence~(GAI) technology has achieved remarkable development across five primary domains: text~\cite{achiam2023gpt,touvron2023llama,liu2024deepseek,li2025predictable}, image~\cite{ho2020denoising,rombach2022high,ramesh2021zero,flux2024,hurst2024gpt,gemini220250312,liu2025step1x}, audio~\cite{wang2023neural,huang2025step,ding2025kimi}, video~\cite{liu2024sora,ma2025step,wan2025}, and 3D content~\cite{mildenhall2021nerf,kerbl20233d,jun2023shap,nichol2022point,hong2023lrm,zhang2024clay}. However, unlike these other modalities, 3D generation technology exhibits significantly lower maturity and slower development progress, remaining far from production-ready. To facilitate further advancements in the field of 3D generation, we analyze the main challenges of 3D content generation from the respective datasets, algorithms, and their corresponding technological ecosystems.

Firstly, data scarcity constitutes the primary bottleneck in 3D generation advancement, where utility is governed by both quantity and quality. Current open-source datasets with >10K samples are limited to ShapeNet~\cite{shapenet2015} (51K), Objaverse~\cite{deitke2023objaverse}~(800K) , and Objaverse-XL~\cite{objaverseXL}(10.2M). Although Objaverse-XL exceeds 10M samples, its web-sourced provenance results in unavoidable quality heterogeneity. Secondly, the inherent complexity of 3D representations—where geometry and texture are decoupled—renders quality assessment fundamentally more challenging than in other modalities. Finally, despite rapid advances, the 3D generation ecosystem remains underdeveloped, with a growing gap between opensource~\cite{li2024craftsman,xiang2024structured,zhao2025hunyuan3d} and proprietary solutions. For instance, an open-source model like Trellis~\cite{xiang2024structured} suffers from limited generalization due to its 500K-scale training dataset. Meanwhile, the release strategy for some advanced models, like Hunyuan3D 2.0~\cite{hunyuan3d22025tencent}, providing only pre-trained weights (without training code), can restrict fine-tuning. 
Furthermore, these models often lack the conditional generation support typically found in commercial platforms like Tripo and Rodin.
These challenges in data availability, reproducibility, and controllability significantly impede progress in 3D generation technologies.

\begin{figure*}[!h]
  \centering
  \includegraphics[width=\linewidth]{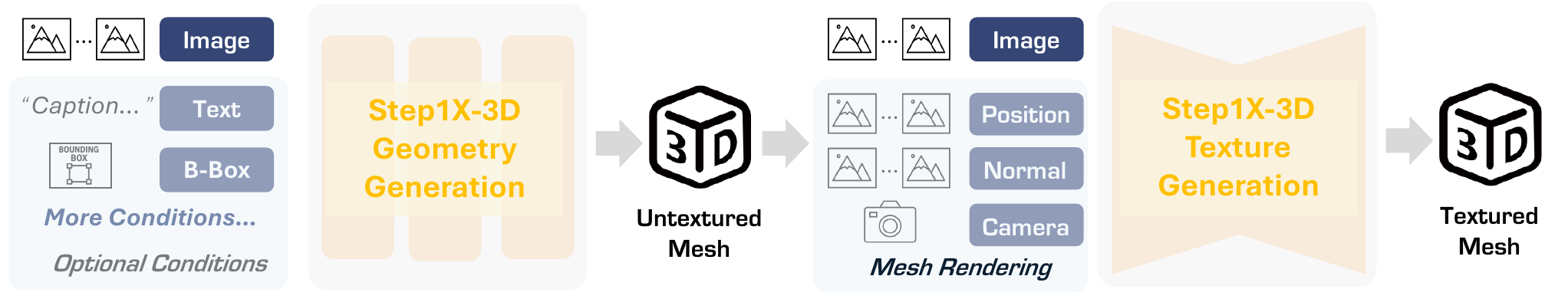}
  \caption{
  \textbf{Overall pipeline of Step1X-3D}.
  }
  \label{fig:overall-framework}
\end{figure*}

To address this disparity, we introduce Step1X-3D, a 3D native framework that combines advanced data curation with an innovative two-stage architecture as shown in Fig.~\ref{fig:overall-framework}.
Our goal is to achieve high-fidelity, controllable 3D asset generation while championing reproducibility through the open release of curated data and training methodologies. 
The Step1X-3D data pipeline begins with over 5M 3D assets sourced from public datasets~(e.g., Objaverse~\cite{deitke2023objaverse}, Objaverse-XL~\cite{objaverseXL}, and etc) and proprietary collections. These assets undergo a rigorous multi-stage refinement process: first, low-quality textures are eliminated based on criteria like resolution, normal map accuracy, material transparency, and surface complexity; second,  watertight mesh conversions are enforced to ensure geometric consistency for robust training supervision. This process yields a curated dataset of 2M high-quality assets, a subset of which (alomost 800K assets derived from public data) will be openly released. 
Architecturally, Step1X-3D employs: (1) a geometry generation stage using a hybrid 3D VAE-DiT diffusion model to produce Truncated Signed Distance Function~(TSDF)~(later meshed via marching cubes), and~(2) a texture synthesis stage leveraging an SD-XL-fine-tuned multi-view generator. This generator conditions on the produced geometry and input images to produce view-consistent textures that are then baked onto the mesh. This integrated approach aims to advance the field by simultaneously resolving critical challenges in data quality, geometric precision, and texture fidelity, while establishing new benchmarks for open, reproducible research in 3D generation.

The Step1X-3D geometry generation framework employs a latent vector set representation~\cite{zhang20233dshape2vecset} to encode point clouds into a compact latent space, which is decoded into TSDF through a scalable perceiver-based encoder-decoder~\cite{jaegle2021perceiver,zhao2023michelangelo} architecture. To preserve high-frequency geometric details, we incorporate sharp edge sampling and integrate dual cross attention mechanisms derived from DoRA~\cite{chen2024dora}. For the diffusion backbone, we adapt the state-of-the-art MMDiT architecture from FLUX~\cite{flux2024} – originally developed for text-to-image generation –by modifying its transformer layers to process our 1D latent space. This VAE-Diffusion hybrid design, architecturally analogous to contemporary 2D generative systems, facilitates direct transfer of 2D parameter-efficient adaptation methods (e.g., LoRA~\cite{ye2023ip}) to 3D mesh synthesis. Consequently, our framework uniquely enables single-view-conditioned pretraining on large-scale datasets while maintaining compatibility with established 2D fine-tuning paradigms for downstream 3D generation tasks, effectively bridging 2D and 3D generative approaches.

The Step1X-3D texture synthesis pipeline initiates with post-processing of the Step1X-3D geometry output using Trimesh to rectify surface artifacts~(including non-watertight meshes, topological irregularities, and surface discontinuities), followed by UV parameterization via xAtlas~\cite{xatlas}. The synthesis process employs a three-stage architecture:~(1) a multi-view image generation diffusion model that conditions on both input images and rendered geometric maps~(normal and position) to enforce view consistency and geometric alignment; (2) a texture-space synchronization module integrated within the denoising process to maintain cross-view coherence through latent space alignment; and (3) texture completion via multi-view back-projection with subsequent texture-space inpainting to address occlusion artifacts and generate seamless UV maps. This hierarchical approach ensures both geometric fidelity and photometric consistency throughout the texture generation pipeline.

In summary, the key contributions of this technical report are threefold:
\begin{itemize}
\item We present a comprehensive data curation pipeline that provides insights into 3D asset characteristics while enhancing generation fidelity.
\item  We propose Step1X-3D, a 3D native generation framework that decouples geometry and texture synthesis. It produces topologically-sound meshes with geometrically-aligned textures, while enabling enhanced controllability through image and semantic inputs. The full framework - including base models, training code, and LoRA-based adaptation modules - will be open-sourced to benefit the 3D research community. 
\item Extensive comparative experiments demonstrate that Step1x-3D surpasses existing open-source 3D generation approaches in asset quality while achieving performance comparable to proprietary state-of-the-art solutions.
\end{itemize}

\section{Related work}
\subsection{Optimization-based 3D Generation}
Optimization-based 3D generation is a framework where 3D representations and their optimization procedures are independently defined across distinct objects, enabling text-to-3D synthesis through input prompts. This approach incorporates diverse 3D representations, ranging from geometric primitives (e.g., voxels, meshes~\cite{botsch2010polygon}) to advanced neural representations such as Neural Radiance Fields~(NeRF)~\cite{mildenhall2021nerf} and 3D Gaussian Splatting~(3D GS)~\cite{kerbl20233d}. Text prompts are encoded into pre-trained models~(e.g., CLIP~\cite{radford2021learning}, Stable Diffusion~\cite{rombach2022high} and etc) to extract semantic-aligned 2D visual features, which supervise the multi-view consistency of rendered 3D representations via differentiable rendering pipelines. By iteratively optimizing geometric and appearance parameters through gradient-based alignment of 2D supervision, the framework achieves cross-modal~(text-to-3D) consistence and produces 3D assets. These optimization-based 3D generation methods~\cite{jain2022zero,sanghi2022clip,poole2023dreamfusion,chen2023fantasia3d,zhuang2025styleme3d}, while capable of processing arbitrary text prompts, suffer from limited support for image-conditioned inputs and inherent limitations such as prolonged optimization cycles, poor generalization of trained parameters and inferior generation quality. These shortcomings critically hinder their practical deployment and scalability in real-world applications.

\subsection{Feed-forward 3D Generation}
To enhance generation efficiency and quality, feed-forward 3D generation methods have emerged as a promising alternative. Specifically, feed-forward methods aim to fit 3D datasets comprising 3D representations~(e.g., multi-view images, point cloud, NeRF, 3D GS and mesh), enabling direct 3D synthesis via forward propagation during inference without iterative optimization, thereby achieving order-of-magnitude acceleration compared to optimization-based approaches. Concurrently, advancements in 3D data availability and algorithmic innovations have driven qualitative leaps in generation fidelity and generalization capability. The evolution of feed-forward 3D generation can be categorized into two distinct paradigms: \textbf{\textit{2D-lifting-to-3D generation paradigm}} and \textbf{\textit{3D native generation paradigm}}.

\paragraph{2D-lifting-to-3D Generation Paradigm}
The 2D-lifting-to-3D paradigm introduces 2D priors into feedforward frameworks, analogous to optimization-based methods, allowing cross-modal knowledge transfer from large-scale 2D datasets. The 2D-lifting-to-3D paradigm is broadly divided into two categories, with the first focusing on single-stage 3D reconstruction from single image input through image embedding techniques (e.g., DINOv2~\cite{oquab2023dinov2}) for feature extraction, followed by 3D geometry synthesis via Vision Transformer~(ViT)\cite{dosovitskiy2020image} architectures. Representative methods include LRM~\cite{hong2023lrm}, which reconstructs Neural Radiance Fields~(NeRF)~\cite{mildenhall2021nerf} from encoded image features, TGS~\cite{Zou_2024_CVPR} that generates 3D Gaussian Splatting~(3D GS)~\cite{kerbl20233d} representations via transformer-based decoders, and TripoSR~\cite{tochilkin2024triposr} and MeshLRM~\cite{wei2024meshlrm}, which leverage hybrid decoders with differentiable rasterization to directly output mesh structures, collectively demonstrating the versatility of this paradigm across diverse 3D representations.
The second category adopts a two-stage 2D-lifting-to-3D paradigm, explicitly prioritizing multi-view consistency for robust 3D generation. These methods accept text prompts or single-image inputs and leverage pre-trained 2D stable diffusion models~\cite{rombach2022high} to synthesize multi-view images, which are subsequently fed into reconstruction networks to produce diverse 3D representations. Research efforts focus on dual optimization: enhancing multi-view synthesis quality in the first stage and improving 3D reconstruction fidelity in the second. For multi-view consistency, SyncDreamer~\cite{liu2024syncdreamer} introduces cost volume constraints to align geometry across views, Wonder3D~\cite{long2024wonder3d} jointly generates RGB images and normal maps for cross-modal consistency, Era3D~\cite{li2024era3d} employs epipolar attention to boost computational efficiency while scaling output resolution from 256×256 to 512×512, and MV-Adapter~\cite{huang2024mv} further extends resolution to 768×768. Beyond 2D image diffusion priors,  SV3D~\cite{voleti2024sv3d} and VideoMV~\cite{zuo2024videomv} leverage stable video diffusion models~\cite{blattmann2023stable} with temporal coherence to achieve superior multi-view alignment. Reconstruction-oriented methods like LGM~\cite{tang2024lgm} and GRM~\cite{xu2024grm} refine neural field representations, while InstantMesh~\cite{xu2024instantmesh}, CRM~\cite{wang2024crm}, and Unique3D~\cite{wu2024unique3d} prioritize practical mesh reconstruction through differentiable rasterization. Approaches including MV-Diffusion++~\cite{tang2024mvdiffusion++}, Meta 3D Gen~\cite{bensadoun2024meta}, and Cycle3D~\cite{tang2025cycle3d} separately enhance algorithmic components across both stages—refining multi-view synthesis pipelines for consistency and optimizing reconstruction networks for geometric fidelity—through meticulous design of each standalone module within the two-phase framework. Since 2d-lifting-to-3d generation methods mainly rely on 2D images as their foundational supervision, they predominantly prioritize the perceptual quality of generated images while neglecting 3D geometric fidelity. Consequently, the resulting 3D geometry frequently exhibits incompleteness and a lack of fine-grained detail.

\paragraph{3D Native Generation Paradigm}
Through explicit geometric representation modeling and learned 3D feature extraction, native 3D generation frameworks enable precise geometry synthesis. In the primitive 3D era, early works like Point-E~\cite{nichol2022point} and Shape-E~\cite{jun2023shap}, constrained by limited 3D datasets~\cite{chang2015shapenet} and underdeveloped architectures, were restricted to limited-category object generation with low visual fidelity. Current native 3D models primarily follow the CG modeling pipeline, decomposing 3D generation into a two-stage process: first generating geometry, then using the geometry to guide texture generation. These native 3D models directly model the geometric component of 3D representations through generative models, achieving more precise and accurate geometric results compared to 2D-lifting-to-3D approaches~\cite{hong2023lrm,tang2024lgm,wang2024crm,long2024wonder3d,xu2024instantmesh,yang2024hunyuan3d}. These significant advancements are largely attributed to recent progress in 3D data and algorithmic developments. On the data side, early native 3D algorithms were primarily trained on ShapeNetCore~\cite{chang2015shapenet}, which contained only about 51,300 objects. Later, with the release of Objaverse~\cite{deitke2023objaverse} and Objaverse-XL~\cite{deitke2023objaverse}, the dataset sizes expanded to 800K+ and 10M+, respectively, greatly enriching the 3D asset database. This played a crucial role in validating model generation capabilities and scalability. Many proprietary algorithms~(e.g., Tripo, Rodin and Hunyuan) have even incorporated additional private data to further enhance generalization or specialized performance. Beyond data, algorithmic advancements have also been pivotal. In 3D Variational Auto-Encoder~(VAE), Michelangelo~\cite{zhao2023michelangelo} innovatively introduced a perceiver-based~\cite{jaegle2021perceiver} architecture for 3D point cloud feature extraction while aligning 3D with text and image modalities. Direct3D~\cite{wu2024direct3d} encodes 3D shaped into a latent triplane space. 3DShape2VecSet~\cite{zhang20233dshape2vecset} proposed a transformer-based method for representing 3D latent space. CLAY~\cite{zhang2024clay} further scale 3DShape2VecSet to a large scale dataset and demonstrate the unprecedented potential of 3D native diffusion model. Trellis~\cite{xiang2024structured} introduced a Structured Latent Representation by aggregating features from 3D voxels and multi-view inputs. Dora~\cite{chen2024dora} proposed a sharp edge sampling strategy to improve geometric detail reconstruction. In generative algorithms, 3D diffusion models have been constructed by adapting 2D generative techniques such as flow matching~\cite{lipman2022flow,liu2022flow} and DiT~\cite{peebles2023scalable} architectures, ensuring robust conditional injection and geometric accuracy. Meanwhile, autoregressive 3D generation methods have also seen rapid development, with some focusing purely on generation (e.g., Mesh-GPT~\cite{siddiqui2024meshgpt}, Mesh-XL~\cite{chen2024meshxl}, LLAMA-mesh~\cite{wang2024llama} and the recent OctGPT~\cite{wei2025octgpt}), while others~(e.g., MeshAnything~\cite{chen2024meshanything}, EdgeRunner~\cite{tang2024edgerunner} and BPT~\cite{weng2024scaling}) optimize dense meshes for intelligent retopology to facilitate downstream tasks like shading and rendering. Currently, mainstream 3D asset generation tools—such as the open-source CraftsMan3D~\cite{li2024craftsman}, Trellis~\cite{xiang2024structured} and Hunyuan3D 2.0~\cite{hunyuan3d22025tencent}, and proprietary solutions like Tripo, Rodin~\cite{zhang2024clay}, and Meshy—primarily rely on 3D diffusion for geometry generation, as autoregressive methods still lag in effectiveness due to limitations in token representation. Beyond geometry generation, texture synthesis has also made significant strides. TEXTure~\cite{richardson2023texture} and Text2Tex~\cite{chen2023text2tex} leverage pre-trained depth-to-image diffusion models~\cite{zhang2023adding}, iteratively generating multi-view images from given camera trajectories and depth maps for texture baking. However, independently generated views suffer from severe consistency issues. Paint-3D~\cite{zeng2024paint3d} adopts a coarse-to-fine approach, introducing UV inpainting and UVHD diffusion models to refine incomplete regions and remove illumination artifacts in UV space. SyncMVD~\cite{liu2024text} enhances multi-view consistency by incorporating a texture synchronization module in latent space, while MVPaint~\cite{cheng2024mvpaint} argues that latent space resolution~($32\times32$) is insufficient for fine texture details and instead performs texture synchronization in image space with resolution~($128\times128$), combined with 3D point cloud inpainting. This technique report follows the 3D diffusion generation paradigm to generate high-fidelity 3D shapes and textures compared to other state-of-the-art~(SOTA) methods, and further involve controllable modules to improve the flexibility of the generation process.

\section{Step1X-3D Geometry Generation}
Step1X-3D is a flow-based diffusion model designed to generate high-fidelity 3D shapes from images, with support for multi-modal conditioning including text and semantic labels. The proposed geometry generation model builds upon prior latent set diffusion models such as Shape2VecSet~\cite{zhang20233dshape2vecset}, CLAY~\cite{zhang2024clay}, Michelangelo~\cite{zhao2023michelangelo}, and Craftsman3D~\cite{li2024craftsman}, utilizing a latent set diffusion framework with rectified flow for 3D shape generation.

In this section, we first introduce the data curation methodology for preprocessing in Sec.~\ref{data-preprocessing}. Next, we provide a detailed description of the architectural design for both the Shape VAE and diffusion model components in Sec.~\ref{shape-generation}. Additionally, inspired by the approach in CLAY~\cite{zhang2024clay}, we present an adaptation of the LoRA~\cite{hu2022lora} ecosystem for 3D generation in Sec.~\ref{controlnet}.
All training code and sampled data will be made publicly available to support research and community development.

\subsection{Geometry Data Curation} 
\label{data-preprocessing}
Recent years have witnessed the emergence of several large-scale open-source 3D datasets, including Objaverse~\cite{deitke2023objaverse}, Objaverse-XL~\cite{objaverseXL}, ABO~\cite{collins2022abo}, 3D-FUTURE~\cite{fu20213d}, ShapeNet~\cite{chang2015shapenet}, etc., which together contain more than 10 million 3D assets. However, as most of this data is sourced from the web-particularly the extensive Objaverse-XL collection-the quality varies considerably. To ensure the data is suitable for training, we implemented a comprehensive 3D data processing pipeline that performs thorough preprocessing to curate a high-quality, large-scale training dataset.

Our curation pipeline consists of three main stages. First, we filter out low-quality data by removing assets with poor textures, incorrect normals, transparent materials or single surface. Second, we convert non-watertight meshes into watertight representations to enable proper geometry supervision. Third, we uniformly sample points on the surface along with their normals to provide comprehensive coverage for the VAE and diffusion model training.
Through our comprehensive data processing pipeline, we successfully curated roughly 2 million high-quality 3D assets from multiple sources: extracting 320k valid samples from the original Objaverse dataset, obtaining an additional 480k from Objaverse-XL, and combining these with carefully selected data from ABO, 3D-FUTURE, and some internal created data.

\begin{figure*}[t!]
  \centering
  \includegraphics[width=\linewidth]{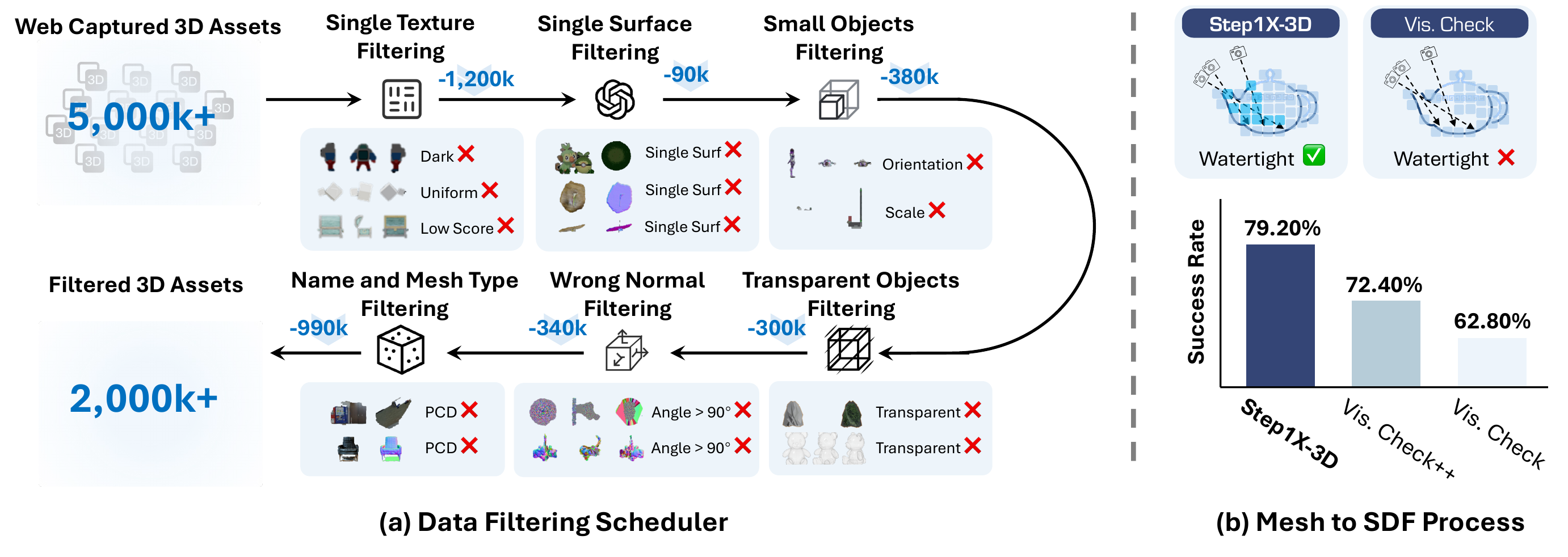}
  \caption{
  \textbf{Data curation pipeline}. (a)~we filter bad quality data from web captured 3D assets (\textit{e.g.}, Objaverse, ABO, Thingiverse, and self-collected assets). (b)~improved conversion rate of mesh-to-SDF.
  }
  \label{fig:winding_number}
\end{figure*}

\paragraph{Data Filter}
The complete data filter process is shown in Fig.~\ref{fig:winding_number}~(a). \textit{(1) Texture Quality Filtering:}
We render 6 canonical-view albedo maps for each 3D model. These rendered images are then converted to HSV color space for analysis. For each view, we compute histograms of the Hue (H) and Value (V) channels. Based on these histograms, we filter out textures that are either too dark, too bright, or overly uniform in color. We then compute perceptual scores for these six views and sort the data accordingly, removing the bottom 20\% of lowest-ranked samples.
\textit{(2) Single-surface Filtering:}
We render 6 canonical-view Canonical Coordinate Maps~(CCM)~\cite{Wang_2019_CVPR, sweetdreamer} to detect single-surface geometry. Specifically, we check whether corresponding pixels on opposite views map to the same 3D point. If the ratio of such pixel matches exceeds a threshold, the object is classified as single-surfaced.
\textit{(3) Small Object Filtering:}
We filter out data where the target object occupies too small an area in frontal views. This occurs in two scenarios: improper object orientation (e.g., a supine human where only feet are visible in front view), or distant objects in multi-object scenes that become too small after normalization. Specifically, we calculate the percentage of valid alpha-channel pixels in frontal views and discard samples with less than 10\% pixels coverage.
\textit{(4) Transparent Object Filtering:}
We exclude objects with transparent materials, as they are typically modeled using alpha-channel planes (e.g., for tree leaves). These transparent surfaces cause misalignment between rendered RGB images and actual geometry, adversely affecting model training. Our filtering method detects and removes assets whose Principled BSDF shaders contain alpha channels.
\textit{(5) Wrong Normal Filtering:}
We identify and remove data with incorrect normals, which would otherwise create holes during watertight conversion. Our method renders 6-view normal maps in camera space and detects erroneous normals by checking whether any normal vectors form obtuse angles with their corresponding camera position.
\textit{(6) Name and Mesh Type Filtering:}
We also filter out data labeled as point clouds by name or mesh type, as these scan-derived datasets typically contain noisy geometry and are difficult to convert into watertight meshes.

\paragraph{Enhanced mesh-to-SDF}
Training a Shape VAE~\cite{zhang20233dshape2vecset, chen2024dora} requires watertight mesh to enable the extraction of SDF (Signed Distance Function) field from the processed meshes, which serve as geometric supervision~\cite{zhang2024clay}.
CLAY~\cite{zhang2024clay} introduced a "visibility check" method for Mesh-to-SDF conversion, which first splits the space into $N^3$ grids. Each grid center checks the visibility using depth test and utilizes a mask with size $N^3$ to indicate whether the grid is considered invisible.
However, for the non-manifold objects with holes, like the windows on the wall, it is easy to encounter floaters inside the converted mesh. 
To address this challenge, we implement a robust classification scheme by incorporating the concept of the winding number~\cite{windingnum} as in CraftsMan3D~\cite{li2024craftsman}, which is an effective tool for determining whether points are inside or outside a shape. 
For each point sampled within the voxel grid, we compute its generalized winding number\footnote{https://libigl.github.io/tutorial/\#generalized-winding-number}, considering points with values above our empirically determined threshold of 0.75. The resulting winding number mask is then combined with the original visibility test through logical conjunction to generate the final occupancy mask for MarchingCubes~\cite{lorensen1998marching}. Experimental results as show in Fig.~\ref{fig:winding_number}~(b) demonstrate this approach achieves a 20\% watertight conversion success rate improvement on the Objaverse dataset~\cite{deitke2023objaverse}.

\paragraph{Training Data Conversion}
(1) Data for VAE: Following the Dora~\cite{chen2024dora}, we employ the Sharp Edge Sampling (SES) strategy to enhance point sampling in geometrically salient regions. Specifically, we combine uniformly sampled points $P_{uniform}$ with additional points $P_a$ sampled from salient areas, forming the final point set $P = P_{uniform} \cup P_{salient}$ along with their corresponding normals, as input to the VAE. For geometry supervision, we sample three distinct sets of points with their SDF values: 200k points within the cube volume, 200k points near the mesh surface with a threshold 0.02, and 200k points directly on the surface.
(2) Data for Diffusion: For training our single-image conditioned flow model, we render each 3D model from 20 randomly sampled viewpoints with camera elevation between -15° and 30°, azimuth between -75° and 75°, and focal lengths randomly selected from orthogonal projection or perspective projections (with focal length uniformly sampled from 35mm to 100mm). We adjust the camera position to ensure the content occupies approximately 90\% of the image. Additionally, we apply common data augmentations such as random flipping (for both images and sampled meshes), color jitter, and random rotations between -10° and 10°.

\subsection{Step1X-3D Shape Generation}
Similar to 2D image generation~\cite{rombach2021highresolution,flux2024}, the Step1X-3D shape generation module consists of a shape autoencoder and a Rectified Flow Transformer. For the sampled point cloud $P$, we first compress it into a 1D tensor using a shape latent set autoencoder~\cite{zhang20233dshape2vecset}, then train the diffusion model with a 1D Rectified Flow Transformer inspired by Flux~\cite{flux2024}. We also support additional components like LoRA  for extended flexibility.

\label{shape-generation}
\paragraph{3D Shape Variational Autoencoder}
The success of the Latent Diffusion Model (LDM)~\cite{rombach2021highresolution} proves that a compact, efficient, and expressive representation is essential for training a diffusion model.
Therefore, we first encode 3D shapes into a latent space and then train a 3D latent diffusion model for 3D generation.
Following the design of 3DShape2VecSet~\cite{zhang20233dshape2vecset}, we adopt a latent vector set representation to encode point clouds into latent space and decode them into geometric functions (e.g., signed distance fields or occupancies). To improve scalability, we adopt a transformer-based encoder-decoder architecture as in recent works~\cite{jaegle2021perceiver,zhao2023michelangelo}. Additionally, we incorporate the Sharp Edge Sampling and Dual Cross Attention techniques proposed in Dora~\cite{chen2024dora} to enhance geometric detail preservation. 
Specifically, we use the downsampled variant of 3DShape2VecSet. Instead of learnable queries, we initialize the latent queries $S = \operatorname{FPS}(P_{uniform}) \cup \operatorname{FPS}(P_{salient}) $ directly with the point cloud itself using the Farthest Points Sampling (FPS). 
We first integrate the information of the concatenated Fourier positional encodings with their respective normals into the shape encoder, forming the actual input to the shape encoder:
$\hat{P} = \operatorname{Concat}(PE(P_c), P_n)$, where $P_c$ is the points position and $P_n$ is the normal. The encoder then processes this input using two cross-attention layers and $L_{e}$ self-attention layers, encoding the points into the latent space via:
\begin{equation}
\begin{aligned}
  \operatorname{Enc}(P) &= 
  \operatorname{SelfAttn}^{(i)}(\operatorname{CrossAttn}(S, \hat{P}_{uniform}), \operatorname{CrossAttn}(S, \hat{P}_{salient})), \\ 
  & \forall{i=1,2,\ldots,L_{e}}. 
\end{aligned}
\end{equation}

Similarly,
we use a perceiver-based decoder that mirrors the architecture of the encoder, and has an extra linear layer $\varphi_{\mathcal{O}}$ to learn to predict the Truncated Signed Distance Function (TSDF) value at $x$:
\begin{equation}
\begin{aligned}
\operatorname{Dec}(x | S) &= \varphi_{\mathcal{O}}(\operatorname{CrossAttn}(PE(x), \operatorname{SelfAttn}^{(i)}(S))), \\
& \forall{i=1,2,\ldots,L_{d}},
\end{aligned}
\end{equation}
where $L_d$ is the number of self-attention layers in the shape decoder. 
Given a query point $x\in\mathbb{R}^3$ in 3D space and a learned latent set $S$, the decoder can output its TSDF value. Then the training objective is as:
\begin{equation}
  \begin{split}
    \hspace*{-\tabcolsep} \mathcal{L}_{VAE} =
     & \mathbb{E}_{x \in \mathbb{R}^3}\left[\operatorname{MSE}\left(\hat{\mathcal{O}}(x | S), \operatorname{Dec}(x)\right)\right] + \lambda_{kl}\mathcal{L}_{kl}, 
  \end{split}
\end{equation}

where $\hat{\mathcal{O}}(x)$ is the ground truth TSDF value of $x$ and the truncated scale is set to $2/256$. The KL divergence loss $\mathcal{L}_{kl}$ is used to regularize the latent space distribution to a standard Gaussian distribution.
Subsequently, we sample query points from a regular grid to obtain their corresponding TSDF values, which are then utilized to reconstruct the final surface using the Marching Cubes~\cite{lorensen1998marching}. We also employ Hierarchical Volume Decoding~\cite{lai2025flashvdm} to accelerate the inference process.
Please refer to the 3DShape2VecSet~\cite{zhang20233dshape2vecset} and Dora~\cite{chen2024dora} for more details.

\begin{figure*}[t!]
  \centering
  \includegraphics[width=\linewidth]{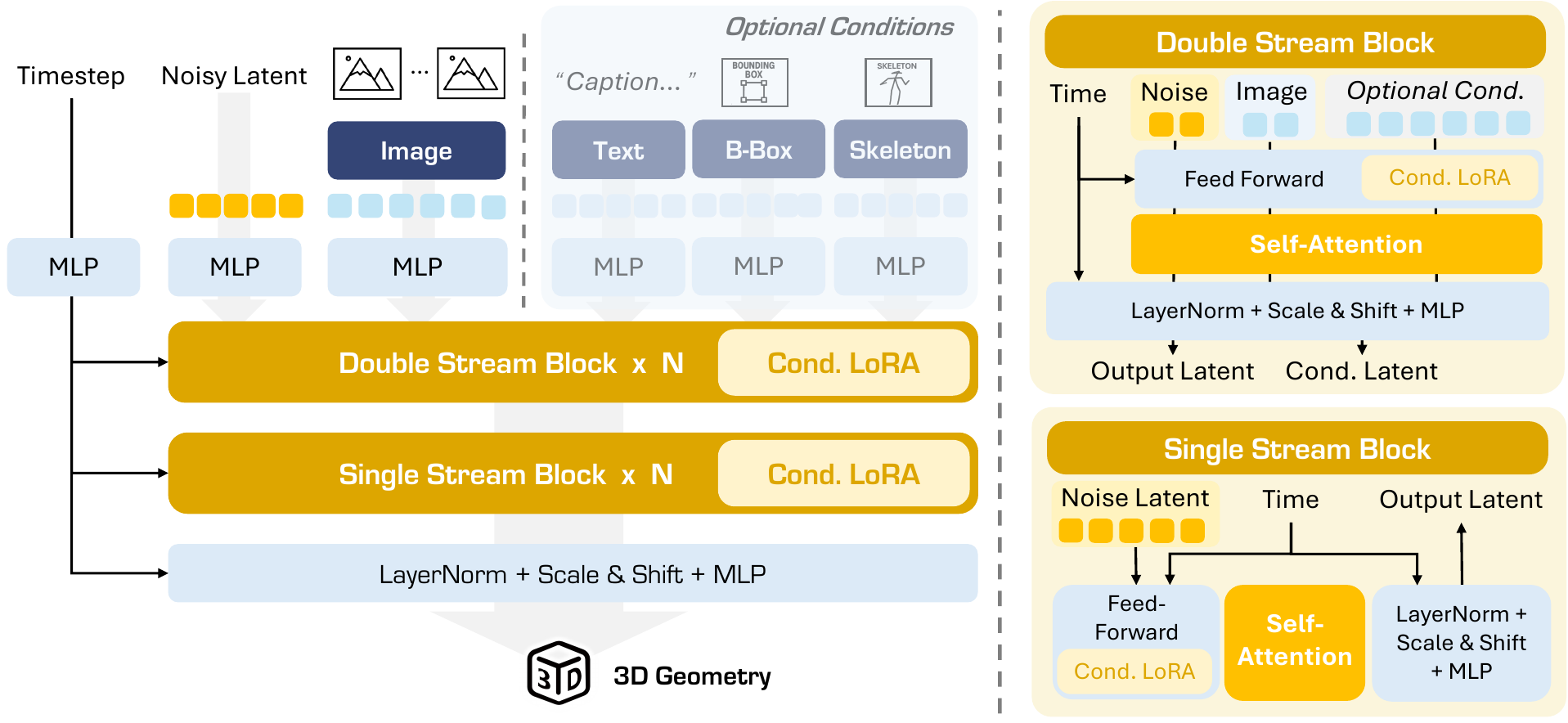}
  \caption{
  \textbf{Step1X-3D geometry diffusion framework}. The framework enables seamless integration of diverse conditions, allowing fine-tuning for various downstream tasks.
  }
  \label{fig:structure}
\end{figure*}

\paragraph{Step1X-3D Diffusion Backbone}
Following the state-of-the-art text-to-image Diffusion
model architectures FLUX~\cite{flux2024}, we use the same MMDiT structure, but modify it for 1D latent space processing as illustrated in Fig.~\ref{fig:structure}. In the dual-stream blocks, latent tokens and condition tokens are processed separately with their own QKV projections and MLPs, but they still interact through cross-attention. In contrast, single-stream blocks combine both types of tokens and process them jointly using parallel spatial and channel attention mechanisms. This hybrid approach allows flexible feature learning while maintaining efficient cross-modal interaction.
Notably, to effectively introduce spatial position information in different noise patches, FLUX.1 employs rotary position encoding (RoPE) to encode spatial
information within each noise patch. Since our ShapeVAE's latent sets representation lacks explicit spatial correspondence, we remove positional embeddings for the latent sets S and retain only timestep embeddings for modulation purposes.
For single-image conditioned shape generation, we leverage a pre-trained DINOv2 large image encoder~\cite{oquab2023dinov2} with registers~\cite{darcet2024visiontransformersneedregisters} to extract conditional tokens from preprocessed 518×518 resolution images - where we perform background removal, object centering/resizing, and white background filling to enhance effective resolution and minimize background interference. 
To capture both semantic and global image characteristics, we concatenate complementary features from CLIP-ViT-L/14~\cite{sanghi2022clip}. These combined features are then injected through parallel cross-attention mechanisms within each flow block, enabling simultaneous processing of both global and local visual information.

\subsection{More flexible control for 3D generation}
\label{controlnet}
Building upon the structural advantages of our VAE with Diffusion framework - which mirrors contemporary text-to-image architectures - we achieve seamless transfer of 2D controllable generation techniques (e.g., ControlNet, IP-Adapter) and parameter-efficient adaptation methods like LoRA to 3D mesh synthesis. As demonstrated by CLAY~\cite{zhang2024clay} in exploring ControlNet-UNet combinations for 3D generation, we systematically implement these control mechanisms within our Step1x-3D framework and support diffusers community~\footnote{https://huggingface.co/docs/diffusers/index}.
To efficiently incorporate conditional signals while preserving the pre-trained model’s generalization ability, we can introduce an additional Condition Branch using the ControlNet-like strategy or LoRA. 
During the current open-source phase, we implement LoRA for geometric shape control using label as reference examples. 
This lightweight solution achieves domain-specific fine-tuning with minimal resource overhead while preserving the original feature space integrity, resulting in controllable shape generation.
We first annotate each mesh with geometric attributes, such as symmetry or level of geometric detail. Using these annotations, we can train an additional LoRA module rather than fine-tuning the entire network, enabling the model to utilize these labels for controlling object geometry.
This LoRA adaptation exclusively to the Condition Branch, enabling efficient injection of conditional signals without compromising the pretrained model's capabilities. 
We will also plan introduce updates in later stages to incorporate fine-tuning through skeleton, bounding box (bbox), caption, and IP-image conditions.

\subsection{Training Rectified Flow Model}
For training, we utilize a flow matching objective that constructs a probability path between Gaussian noise $\mathcal{N}(\mathbf{0}, \mathbf{I})$ and data distributions, where Rectified Flow's linear sampling mechanism streamlines network training by directly predicting the velocity field $\mathbf{u}_t = \frac{d\mathbf{x}_t}{dt}$ that transports samples $\mathbf{x}_t$ toward target data $\mathbf{x}_1$, thereby improving both efficiency and training stability. Building on SD3's logit-normal sampling strategy, we strategically increase the sampling weight for intermediate timesteps $t \in (0,1)$ during training, as these mid-range temporal coordinates pose greater prediction challenges for velocity estimation in the Rectified Flow framework. The final objective is formulated as:

\begin{equation}
\mathcal{L} = \mathbb{E}_{t,\mathbf{x}_0,\mathbf{x}_1}\left[\| \mathbf{u}_\theta(\mathbf{x}_t, \mathbf{c}, t) - \mathbf{u}_t \|^2_2 \right]
\end{equation}

where $\mathbf{c}$ denotes the conditioning signal, with an adaptive time-step weighting scheme. To further stabilize training, we incorporate an Exponential Moving Average (EMA) strategy with a decay rate of 0.999 to smooth parameter updates. The training is conducted in two phases: initially, for rapid convergence, we use a latent set size of 512, a learning rate of 1e-4, and a batch size of 1920 across 96 NVIDIA A800 GPUs for 100k iterations. Subsequently, to enhance model capacity and precision, we scale the latent set size to 2048, reduce the learning rate to 5e-5, and halve the batch size to 960 for another 100k iterations, ensuring robust adaptation to higher-dimensional data spaces while maintaining computational efficiency.

\section{Step1X-3D Texture Generation}

\begin{figure*}[!h]
  \centering
  \includegraphics[width=\linewidth]{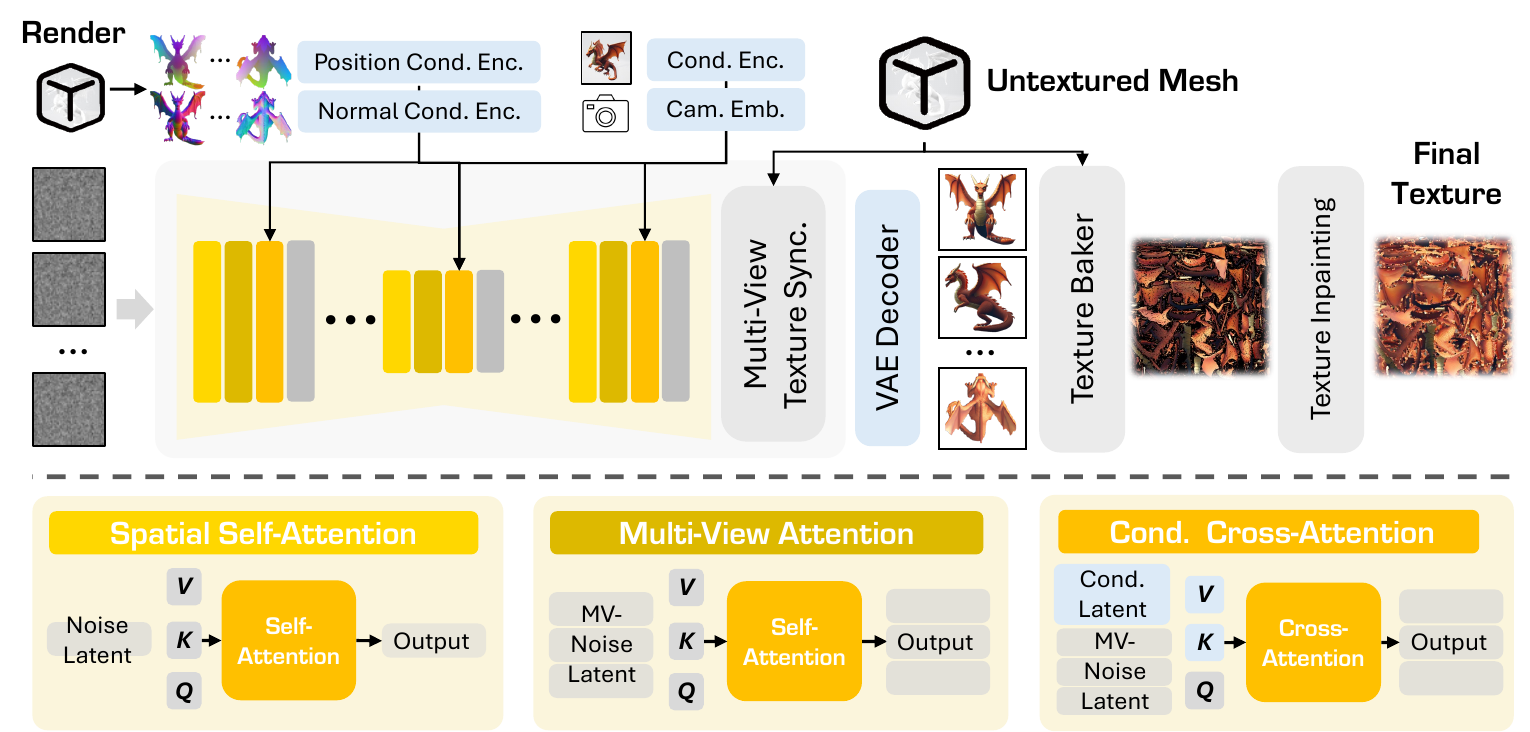}
  \caption{
  \textbf{Texturing module of Step1X-3D}.
  }
  \label{fig:texture-framework}
\end{figure*}

Once the untextured 3D geometry is generated using the Step1X-3D framework, texture synthesis is performed via a multi-stage pipeline, as shown in Fig.~\ref{fig:texture-framework}. 
First, the raw geometry undergoes post-processing to ensure topological consistency and structural integrity~(Sec.~\ref{texture:pose_process}). Then, we perpare 3D assets for texture generation~(Sec.~\ref{texture:texture_dataset}). Next, a multi-view image generation model is fine-tuned on high-quality 3D datasets, incorporating geometric guidance via normal and position maps~(Sec.~\ref{texture:mv_gen}).
Finally, the generated multi-view images are super-resolved to 2048×2048 resolution before UV baking, followed by inpainting to complete the texture maps~(Sec.~\ref{texture:baker}).

\subsection{Geometry Postprocess}
\label{texture:pose_process}
To achieve high-fidelity texturing, we perform post-processing on the mesh geometry generated by the preceding geometric generation pipeline. The optimization process primarily employs the trimesh toolkit~\cite{trimesh}. Specifically, we first verify the watertightness of the initial mesh, implementing hole-filling algorithms where non-manifold geometry is detected. Subsequently, we apply a remeshing operation that subdivides each triangular face into four sub-faces while enforcing Laplacian surface smoothing constraints. This remeshing procedure ensures uniform topological distribution and minimizes UV seam artifacts. Finally, we utilize the xAtlas~\cite{xatlas} parameterization framework to generate optimized UV coordinates, which are then integrated into the final mesh representation. This systematic refinement pipeline guarantees geometric robustness for subsequent texture mapping.

\subsection{Texture datatset preparation}
\label{texture:texture_dataset}
Compared to geometry generation, the texture generation component does not require millions of training samples but instead places higher demands on texture quality and aesthetic metrics. From the 320K Objaverse dataset cleaned in Sec.~\ref{data-preprocessing}, we further curated 30K 3D assets for multi-view generative model training. Specifically, we rendered each object with blender to produce six views (front, back, left, right, top, and bottom), along with corresponding albedo, normal map, and position map outputs with $768\times768$ resolution.

\subsection{Geometry-guided Multi-view Images Generation}
\label{texture:mv_gen}
\paragraph{Single View to Multi-view Generation} 
Given a single view image and taget muliti-view camera poses~(the condtions are defined as $\mathcal{C}$), we aim to generate consistent mulit-view images with a diffusion model $D_{MV}$. In details, we formulate the single-view to multi-view diffusion process as:
    \begin{equation}
        I_{1,2,...,N} = D_{MV}(z^{MV}, \mathcal{C}),
    \end{equation}
where $z^{MV}$ is a multi-view random noise. 
We use the pre-trained MV-Adapter~\cite{huang2024mv} as the backbone to generate multi-view images with $768\times768$ resolution and higher consistency. MV-Adapter exhibits two distinct advantages: the capability to generate high-resolution images and enhanced generalization performance. The high-resolution generation is primarily achieved through a memory-efficient epipolar-attention, which enables the production of 768×768 resolution images under batch size constraints during the training process. The superior generalization capability stems from preserving the original spatial self-attention parameters of SD-XL~\cite{rombach2022high}, while introducing a triple parallelized attention architecture that simultaneously addresses generalization capacity, multi-view consistency, and conditional adherence. This design achieves an optimal balance between maintaining foundational model properties and acquiring specialized generation capabilities.

\paragraph{Involve Geometry Guidance for Generation}
Our objective is to generate reasonable and refined textures for the aforementioned untextured 3D meshes. To achieve this, during multi-view generation, in addition to conditioning on the provided single-view input, the injection of geometric information facilitates enhanced detail synthesis and improved alignment between textures and the underlying mesh surface. Specifically, we introduce two types of geometric guidance: normal maps and 3D position maps. The normal maps preserve fine-grained geometric details of the object, while the 3D position maps ensure accurate spatial correspondence between textures and mesh vertices across different viewpoints through 3D coordinate consistency. Both geometric representations are derived in the global world coordinate system. These geometric features are encoded via image-based encoders and subsequently injected into the backbone generation model through cross-attention mechanisms. This approach enables explicit geometric conditioning while maintaining the generative model’s ability to synthesize perceptually coherent textures.

\paragraph{Synchronized Multi-view Images in Texture Domain}
While the integration of cross-view attention and two geometric conditioning terms has achieved satisfactory multi-view consistency, inherent discrepancies between image space and UV space still introduce artifacts in synthesized textures, such as localized blurriness and discontinuous seams. To address this, we extend the MV-Adapter framework by introducing a texture-space synchronization module during inference. Unlike text-to-multi-view approaches like MVPaint~\cite{cheng2024mvpaint} and SyncMVD~\cite{liu2024text} — which bypass explicit modeling of style references~(sref) and content references (cref) between the input condition image and output multi-view images—our method eliminates the need for auxiliary refinement pipelines~(e.g., Satble diffuson with controlNet) for multi-view synchronization. This design choice is justified by two considerations: 1) Our generator operates at a latent resolution of $96\times96$, which empirically provides sufficient texture representation capacity; 2) Joint optimization in a unified latent space inherently preserves texture coherence across views. Consequently, we implement texture synchronization exclusively through latent space alignment within a single diffusion model backbone, achieving parameter efficiency while maintaining visual fidelity. \\
In details, to predict the multi-view latent output $z_{t+1}$, we unproject the latent $z_{i,t}$ of each view to texture space and $\mathcal{P}$ represents UV mapping function. Then, we obtain the synchronized texture $T_{i,t}^{'}$ by fusing multi-vew $T_{i,t}$ and weight different views in texture space with the cosine similarities between the ray direction and per-pixel normal map. Further, the synchronized latent $z_{i,t}^{'}$ is obtained by projecting the $T_{i,t}^{'}$ with UV rasterization function $\mathcal{P}$. We formulate the whole procession of each denoise step as the following:
\begin{gather}
    T_{i,t} = \mathcal{P}^{-1}(z_{i,t}), \\
    T_{i,t}^{'} = \sum_{i=1}^{N} cos(v_{i}, n_{i})*T_{i,t}, \\
    z_{i,t}^{'} = \mathcal{P}(T_{i,t}^{'}), \\
    z_{t+1} = D_{MV}(z_{t}^{'}, \mathcal{C}).
\end{gather}

\subsection{Bake Texture}
\label{texture:baker}
Following conventional texture baking workflows~\cite{zeng2024paint3d,cheng2024mvpaint,zhao2025hunyuan3d}, we adopt standard texture processing operations on multi-view projections of the object and reuse the texture baker tools in Hunyuan3D 2.0~\cite{zhao2025hunyuan3d}. First, multi-view images were upsampled to achieve a $2048\times2048$ resolution and then are inversely projected onto the texture space. Due to occlusions and multi-view inconsistencies, this process inevitably introduces artifacts such as discontinuities and holes in the UV-mapped texture. To address this, we implement continuity-aware texture inpainting through iterative optimization, ensuring seamless texture synthesis across the entire surface. This post-processing stage effectively resolves topological ambiguities while preserving high-frequency texture details critical for photorealistic rendering.

\section{Experiment}
This section presents a comprehensive evaluation of the generation performance of Step1X-3D. First, we provide a detailed demonstration of Step1X-3D's ability to generate geometry and texture conditioned on a single input image in Sec.~\ref{exp:visual_quality}. Next, we validate the model’s flexibility and controllability in Sec.~\ref{exp:control_res}. Finally, we conduct a thorough comparison between Step1X-3D and state-of-the-art~(SOTA) methods, including both open-source~(Trellis~\cite{xiang2024structured}, Hunyuan3D 2.0~\cite{hunyuan3d22025tencent}, and TripoSG~\cite{li2025triposg}) and proprietary approaches~(Meshy-4~\footnote{The API was invoked from Meshy platform in May 2025.}, Tripo-v2.5~\footnote{The API was invoked from Tripo platform in May 2025.}, and Rodin-v1.5~\footnote{The API was invoked from Rodin platform in May 2025.}), across three key dimensions: quantitative metrics, user studies, and visual quality in Sec.~\ref{exp:sota_compare}.
\subsection{The Visual Quality Results of Step1X-3D Assets}
\label{exp:visual_quality}
\begin{figure*}[!h]
  \centering
  \includegraphics[width=\linewidth]{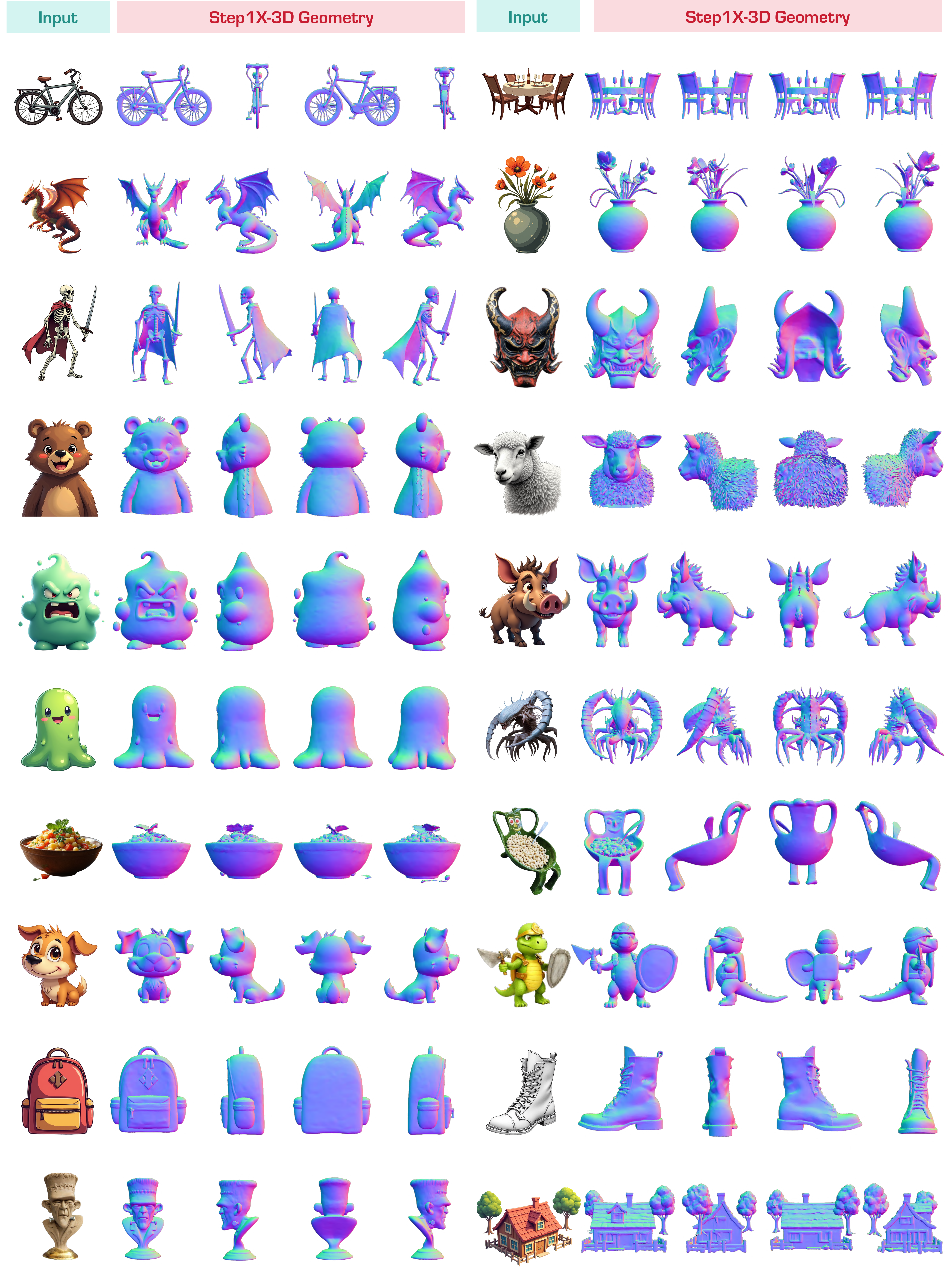}
  \caption{
  \textbf{Samples of generated geometry by Step1X-3D}.
  }
  \label{fig:step1x-3d-gallery-geometry}
\end{figure*}

\begin{figure*}[!h]
  \centering
  \includegraphics[width=\linewidth]{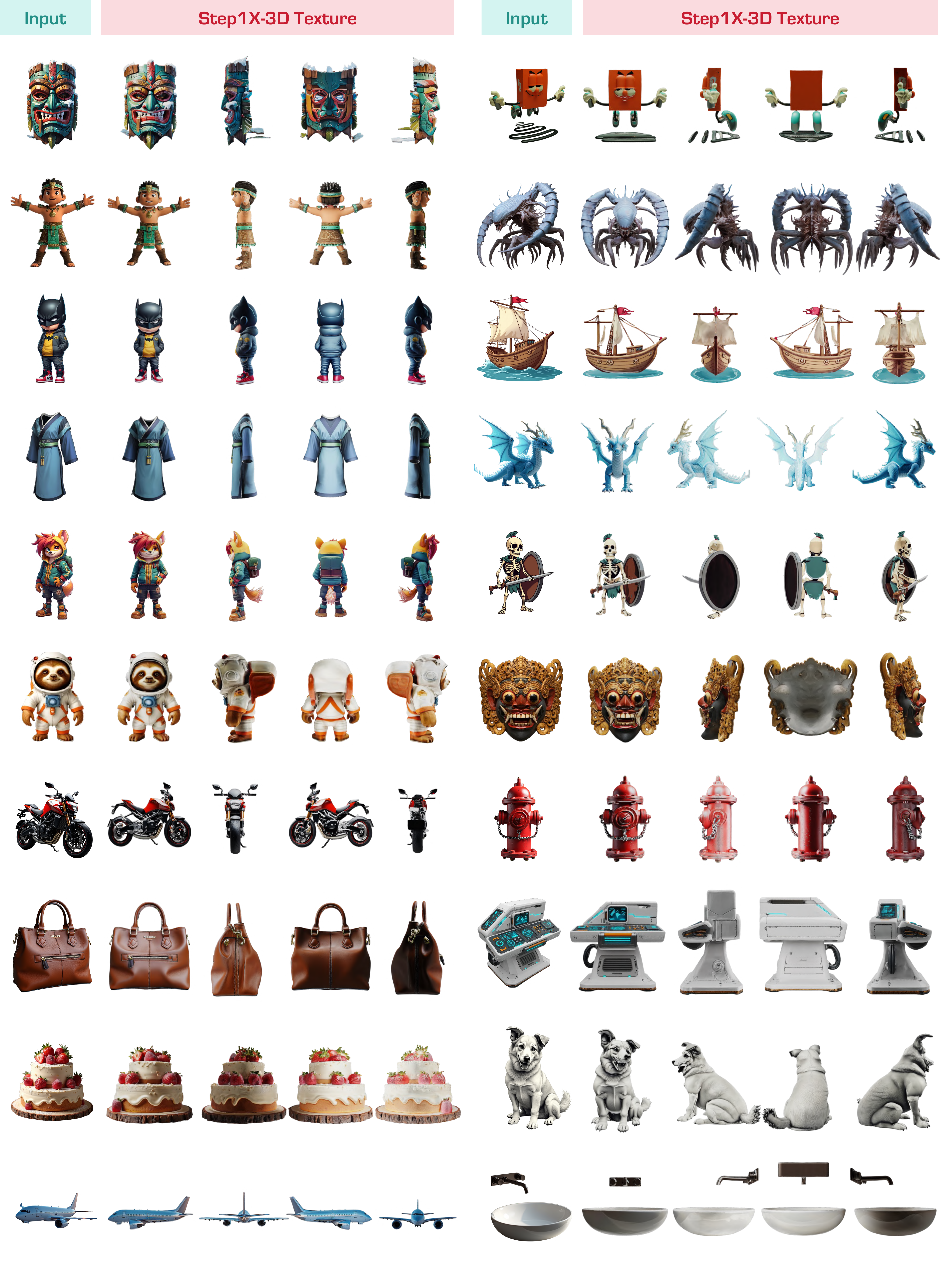}
  \caption{
  \textbf{Samples of generated texture by Step1X-3D}.
  }
  \label{fig:step1x-3d-gallery-texture}
\end{figure*}

To evaluate Step1X-3D, we present the generated 3D assets from both geometric and textural dimensions in Fig.~\ref{fig:step1x-3d-gallery-geometry} and Fig.~\ref{fig:step1x-3d-gallery-texture}, respectively. To better showcase geometric details, we render multi-view normal maps from the generated meshes for 3D geometry visualization. As shown in Fig.~\ref{fig:step1x-3d-gallery-geometry}, the first and sixth columns display input images, while the remaining columns present multi-view representations of different objects. Our test objects cover a wide variety of styles (cartoon, sketch, and photorealistic), geometric complexity (flat surfaces, hollow structures, and detail-rich objects), and spatial configurations (single objects and multi-object compositions). Across this diverse input images, Step1X-3D geometry generation model not only maintains strong similarity between the 3D mesh and input image, but also reconstructs plausible spatial structures for occluded regions with reasonable geometric details. These results demonstrate the crucial role of our specifically designed 3D diffusion model and VAE architecture in Step1X-3D, along with the significant improvement in generalization capability enabled by large-scale high-quality training data. Fig.~\ref{fig:step1x-3d-gallery-texture} further demonstrates Step1X-3D's texture generation capability through multi-view renders of textured 3D meshes. The texture generation model produces style-consistent textures across various input styles while maintaining high fidelity to the input image's textural details. For occluded regions in the input image, by preserving the original SD-XL parameters and incorporating the target model's normal maps and position maps as geometric guidance, Step1X-3D achieves plausible view completion with excellent multi-view consistency and precise geometry-texture alignment. In summary, Step1X-3D generates geometrically plausible 3D geometry with rich textures, where the final textured 3D meshes exhibit strong content and style matching with the input conditioning images.

\clearpage

\begin{figure*}[t!]
  \centering
  \includegraphics[width=\linewidth]{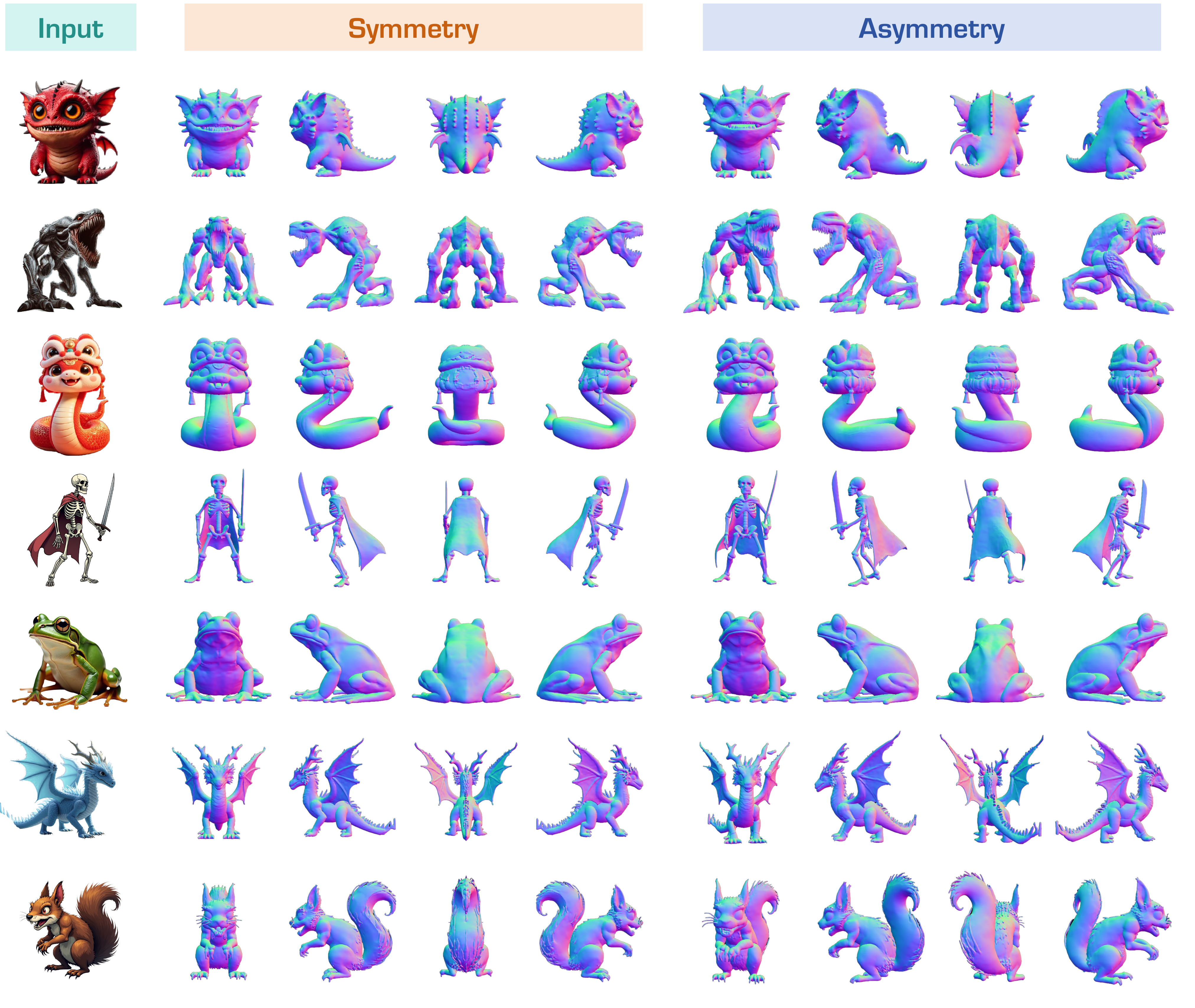}
  \caption{
  \textbf{Symmetry control}. Controllable 3D generation with symmetry or asymmetry geometry.
  }
  \label{fig:symmetry}
\end{figure*}

\begin{figure*}[htbp]
  \centering
  \includegraphics[width=\linewidth]{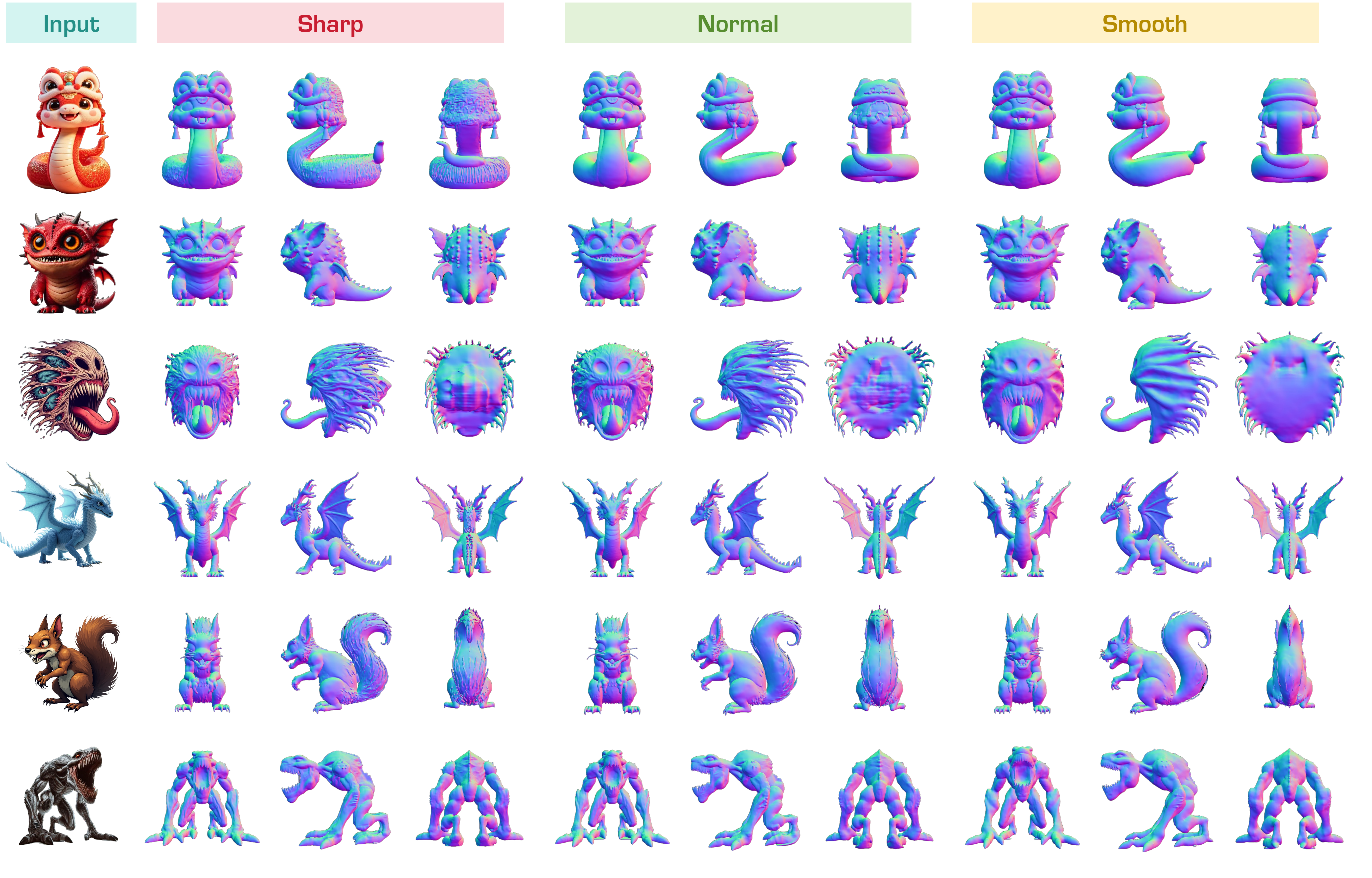}
  \caption{
  \textbf{Sharpness control}. Controllable 3D generation with sharp, normal or smooth geometry.
  }
  \label{fig:detail}
\end{figure*}
\clearpage
\subsection{High Controllable 3D generation with LoRA Finetuning}
\label{exp:control_res}
Building upon a pre-trained geometry generation model for mesh reconstruction conditioned on a single image, we seamlessly integrate LoRA fine-tuning to enable flexible control over diverse 3D generation models. In this experiment, focusing on practical user applications, we specifically design two geometric control strategies: symmetry manipulation and hierarchical geometric detail adjustment. To implement these controls, we collected approximately 30,000 3D models and utilized the Step1O multimodal model to annotate each object based on (1) symmetry properties and (2) geometric detail levels (sharp, normal, and smooth). We show the high controllable 3D generation results in Fig.~\ref{fig:symmetry} and Fig.~\ref{fig:detail}. To better capture geometric details in 3D meshes, we employ multi-view normal maps for geometric representation. Fig.~\ref{fig:symmetry} shows the results of geometry generation using "symmetry"/"asymmetry" captions. The first column shows the input image, columns 2–5 display four views~(front, back, left, right) of the 3D object generated with a symmetry-conditioned caption, while columns 6–9 present the corresponding multi-view results for the asymmetry-conditioned generation. The results indicate that the generated 3D objects consistently adhere to their respective control instructions, with particularly pronounced compliance in the front and back views. Fig.~\ref{fig:detail} illustrates the hierarchical control of geometric details in detail. From left to right, we present the input conditioning image, followed by objects generated with the "sharp", "normal", and "smooth" labels respectively. Each object is represented using normal maps from front, right, and back views. Consistent with previous results, the generated objects demonstrate strong adherence to their corresponding control labels. This further validates the efficacy of Step1X-3D’s fine-tuning technique and demonstrates strong generalization capability in its geometry generation model.

\subsection{Comparison Results with SOTA Methods}
\label{exp:sota_compare}
To further validate the effectiveness of Step1X-3D, we conducted comprehensive comparisons with existing state-of-the-art (SOTA) methods, including open-source approaches (Trellis~\cite{xiang2024structured}, Hunyuan3D 2.0~\cite{hunyuan3d22025tencent}, and TripoSG~\cite{li2025triposg}) and proprietary systems (Tripo-v2.5, Rodin-v1.5, and Meshy-4). Specifically, we performed:~(1) quantitative evaluations using both geometric and textural metrics; (2) user studies assessing perceived 3D quality through subjective scoring; (3) and visual comparisons of geometric and textural results across diverse input conditions.

\paragraph{Quantity Comparisons and User Study across SOTA Methods} Beyond visual comparisons under diverse input conditions, we constructed a comprehensive benchmark dataset totaling 110 images in the wild. This benchmark incorporates: (1) example images from various 3D generation platforms~(e.g., Tripo, Rodin and etc.), and (2) images generated by the Flux model covering 80 object categories from the COCO~\cite{lin2014microsoft} dataset. Based on this test set, we systematically collected 3D assets generated by different methods for both quantitative evaluation and subjective user study.
\begin{table}[!htbp]
\centering
\setlength{\tabcolsep}{2.0pt}
\caption{\textbf{Quantitative comparison on different models}. * indicates closed-source models, and \legendsquare{colorbest}~\legendsquare{colorsecond}~\legendsquare{colorthird} indicate the best, second best, and third best performance respectively.}

\begin{tabular}{l @{\hspace{10pt}}|@{\hspace{10pt}}c@{\hspace{10pt}}|@{\hspace{10pt}} c@{\hspace{10pt}}c@{\hspace{10pt}}c}
\toprule
% \hline
\multirow{2}{*}{\textbf{Model}} & \textbf{Texture} & \multicolumn{3}{c}{\textbf{Geometry}} \\
% \cmidrule{2-5}
% \hline{2-5}
&\textbf{CLIP-Score$\uparrow$} & \textbf{Uni3D-I$\uparrow$} & \textbf{OpenShape$_{sc}$-I$\uparrow$} &
\textbf{OpenShape$_{pb}$-I$\uparrow$}\\

\midrule
% \hline
Rodin-v1.5$^*$ & 0.845 & 0.270 & 0.104 & 0.081\\
Meshy-4$^*$ & 0.796 & \third{0.357} & \third{0.137}& 0.099\\
Tripo-v2.5$^*$ & \second{0.848} & \first{\textbf{0.366}} & \first{\textbf{0.140}}& \second{0.134} \\
\midrule
% \hline
Trellis & \third{0.848} & 0.353 & 0.136 & \first{\textbf{0.139}}\\
Hunyuan3D 2.0 & 0.829 & 0.352 & 0.131 & \third{0.131} \\
\textbf{Step1X-3D~(Ours)} & \first{\textbf{0.853}} & \second{0.361}& \second{0.139} &0.130\\
\bottomrule
% \hline
\end{tabular}
\label{tab:model_comparison}
\end{table}

We similarly designed quantitative metrics for both geometry and texture dimensions. For geometric evaluation, we leverage self-supervised multimodal models to perform feature matching between input 2D images and generated 3D point clouds (extracted from the output meshes). To ensure comprehensive and fair comparison, we utilized two distinct multimodal frameworks for feature extraction: Uni3D~\cite{zhou2023uni3d} and OpenShape~\cite{liu2023openshape}, with cosine similarity serving as our similarity metric. For the OpenShape framework, which follows a self-supervised paradigm, we implemented both SparseConv~\cite{choy20194d} and PointBERT~\cite{yu2022point} as backbone architectures. This yielded three distinct metrics for evaluating image-to-geometry alignment:~Uni3D-I, OpenShape$_{sc}$-I, and OpenShape$_{pb}$-I, where higher scores indicate better geometric consistency with the input image. For texture evaluation, we adopted CLIP-Score~\cite{radford2021learning} to measure semantic alignment. Specifically, we rendered multi-view images from textured 3D models at an elevation of 30° and azimuth angles of \{0°, 90°, 180°, 270°\} for semantic consistency assessment with input images. Quantitative results are presented in Tab.~\ref{tab:model_comparison}, with top and second-highest scores highlighted. Step1X-3D achieved the highest CLIP-Score and multiple second-highest rankings in geometric-semantic matching metrics. These superior results further demonstrate Step1X-3D's robust generation capabilities.

\begin{wrapfigure}{r}{0.55\textwidth}
  \vspace*{-12pt}
  \begin{center}
    \includegraphics[width=0.5\textwidth]{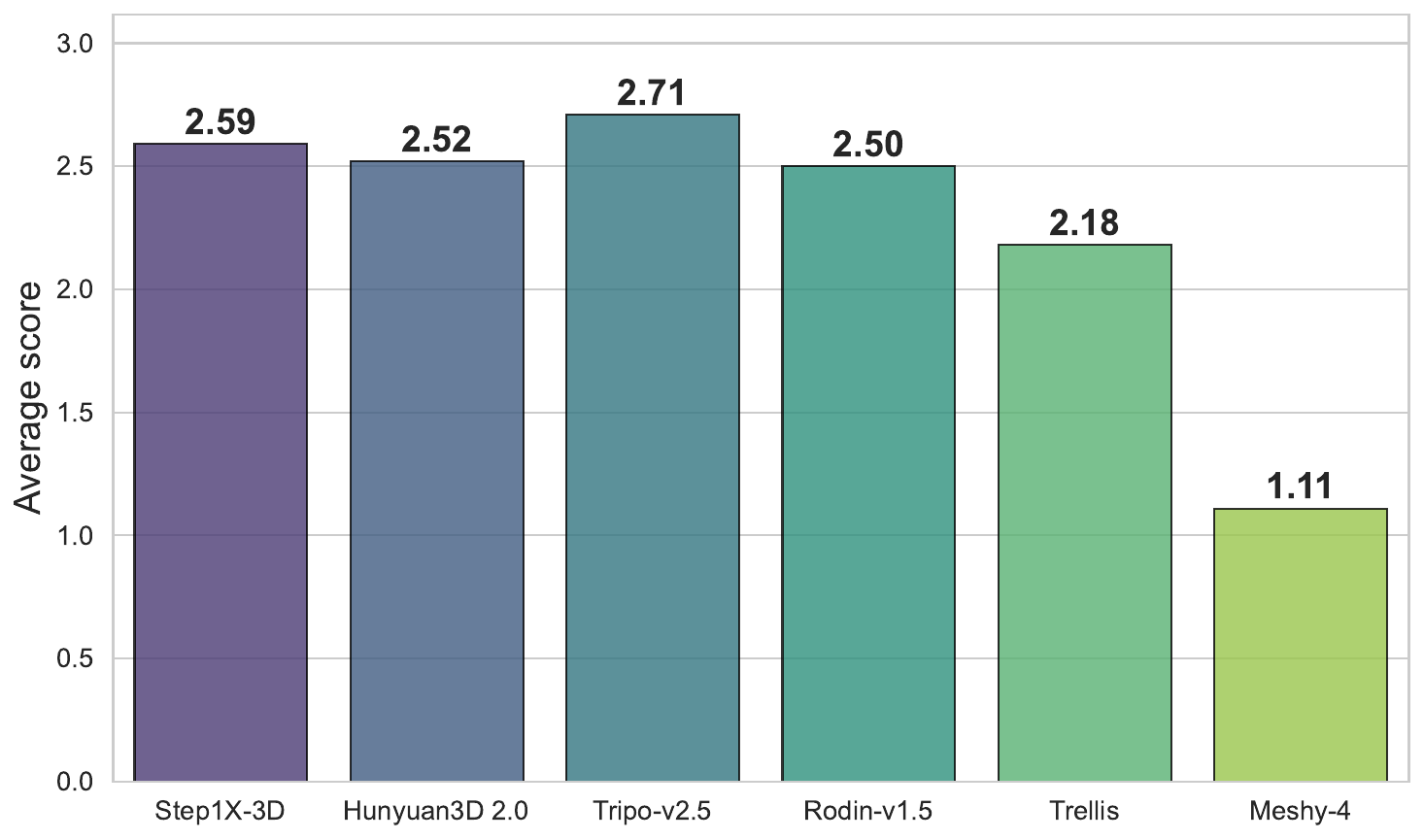}
  \end{center}
  % \vspace*{-20pt}
  \caption{\textbf{User study}. Average user preference score is reported.}
  \label{fig:user_study}
\end{wrapfigure}
We conducted a user study with 20 participants evaluating all 110 uncurated test images. The assessment criteria for 3D models included: (1) geometric plausibility, (2) similarity to input images, (3) texture clarity, and (4) texture-geometry alignment. Participants rated each object on a 5-point Likert scale (1: lowest quality, 5: highest quality). As shown in Fig.~\ref{fig:user_study}, Step1X-3D achieves comparable performance to the current best-performing methods. However, we observe that all evaluated algorithms still fall significantly short of the theoretical upper bound, indicating considerable room for improvement before reaching production-ready quality. These findings underscore the importance of fostering open-source collaboration within the 3D research community to collectively advance the state-of-the-art.

\begin{figure*}[h]
  \centering
  \includegraphics[width=\linewidth]{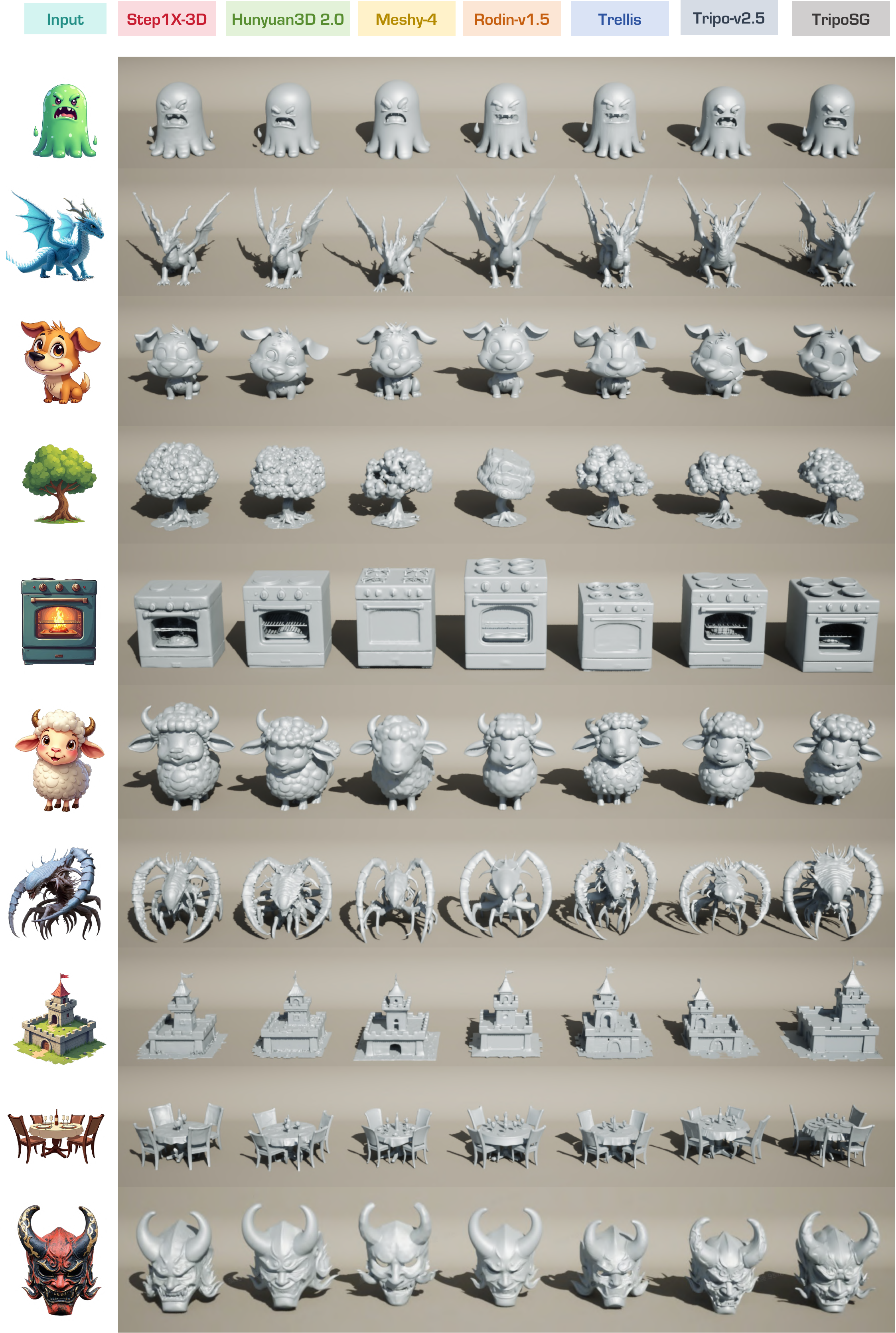}
  \caption{
  \textbf{Qualitative comparison with SOTA methods on generated geometry}.
  }
  \label{fig:comparison-geometry}
\end{figure*}

\begin{figure*}[!h]
  \centering
  \includegraphics[width=0.9\linewidth]{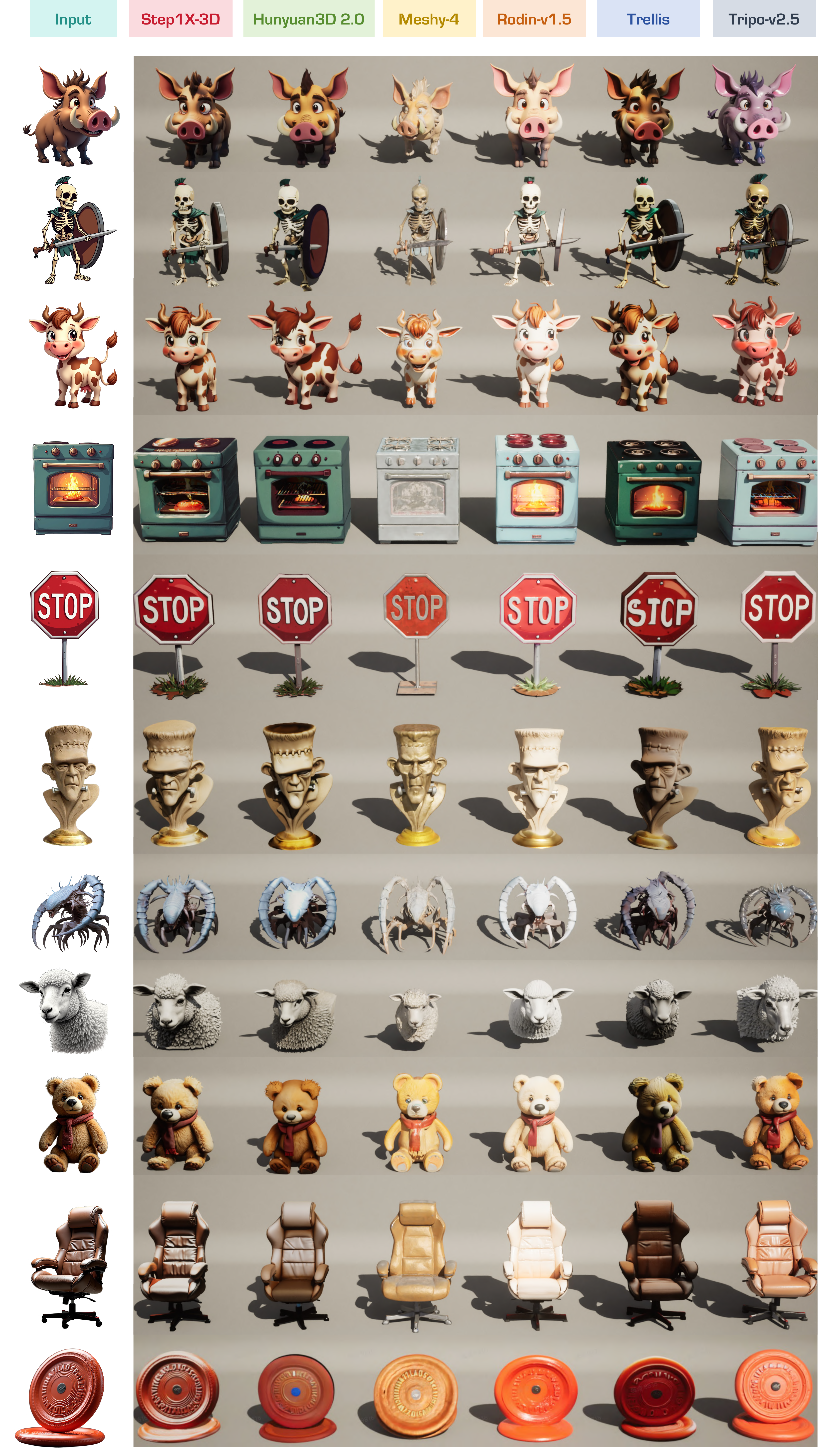}
  \caption{
  \textbf{Qualitative comparison with SOTA methods on generated texture}.
  }
  \label{fig:comparison-texture}
\end{figure*}

\vspace{-0.3cm}
\paragraph{Visual Comparisons across SOTA Methods} Fig.~\ref{fig:comparison-geometry} and Fig.~\ref{fig:comparison-texture} present comparative results of geometric and textural outputs across different methods. Unlike previous visual comparisons, we address pose inconsistencies in the generated 3D meshes by implementing a unified evaluation protocol: (1) aligning both untextured and textured models in Unreal Engine for consistent pose normalization, and (2) compositing multiple objects into a single rendered image for direct comparison. This standardized approach reveals that Step1X-3D achieves comparable or superior performance relative to the best available methods.
\vspace{-0.2cm}

\section{Conclusion}
\vspace{-0.3cm}
Step1X-3D advances 3D generation by introducing an open-source, high-fidelity framework that decouples geometry and texture synthesis. Through rigorous data curation (2M assets) and a hybrid VAE-DiT architecture, it achieves promising results while enabling 2D-to-3D control transfer. We will release the models, training code, as well as the training data~(excluding self-collected assets) to bridge the gap between proprietary and open research, fostering community progress toward production-ready 3D generation.
\vspace{-0.3cm}

\section{Limitations}
\vspace{-0.3cm}
Currently, we convert mesh to TSDF with grid resolution $256^{3}$. In future work, we will increase the grid resolution to achieve more accurate geometric details. Meanwhile, for the texture component, our current implementation is limited to albedo generation. We plan to extend this pipeline to support input image relighting and physically based rendering (PBR) material texture generation.
\vspace{-0.3cm}

\section{Contributors}
\vspace{-0.2cm}
\textbf{Algorithm contributors:}~Weiyu Li, Xuanyang Zhang, Zheng Sun, Di Qi, Hao Li, Wei Cheng, Weiwei Cai, Zeming Li, Gang Yu, Xiangyu Zhang, Daxin Jiang 

\textbf{Data contributors:}~Shihao Wu, Jiarui Liu, Zihao Wang, Xiao Chen, Feipeng Tian, Jianxiong Pan

\textbf{Corresponding authors:}~Xuanyang Zhang~(zhangxuanyang@stepfun.com), Gang Yu~(yugang@step\\fun.com), Daxin Jiang~(djiang@stepfun.com), Ping Tan~(pingtan@ust.hk)

\clearpage
{
    \small
    \bibliographystyle{plain}
    \bibliography{main}
}
\end{document}